\documentclass[sigconf, 10pt]{acmart}

\usepackage{hyperref}

\setcopyright{none}
\renewcommand\footnotetextcopyrightpermission[1]{}
\usepackage{soul}

\usepackage{multirow}
\usepackage[utf8]{inputenc}
\usepackage[ruled, lined, longend, linesnumbered]{algorithm2e}
\usepackage{amsmath,multicol}
\usepackage{subcaption,textcomp,comment}
\usepackage{threeparttable}

\usepackage{booktabs}
\usepackage{multirow}

\usepackage{array}
\newcolumntype{L}[1]{>{\raggedright\let\newline\\\arraybackslash\hspace{0pt}}m{#1}}
\newcolumntype{C}[1]{>{\centering\let\newline\\\arraybackslash\hspace{0pt}}m{#1}}
\newcolumntype{R}[1]{>{\raggedleft\let\newline\\\arraybackslash\hspace{0pt}}m{#1}}

\newcommand{\sys}[0]{\texttt{FedRDMA}\xspace}
\newcommand{\syse}[0]{\texttt{FedRDMA-E}\xspace}

\usepackage{wrapfig}
\usepackage{comment}
\usepackage{colortbl}

\usepackage{color,soul}
\usepackage{graphics}
\usepackage{hyperref}
\usepackage{lipsum}
\usepackage{tikz}
\usepackage{enumitem}	
\usepackage{listings} 

\usepackage{booktabs}	
\usepackage{lipsum}

\usepackage{capt-of}	

\usepackage{todonotes}  
\usepackage{subcaption} 

\definecolor{refkey}{rgb}{0,0,1}
\definecolor{labelkey}{rgb}{0,0,1}

\usepackage{cleveref}
\AtBeginDocument{\DeclareCaptionSubType{lstlisting}}
\crefname{sublstlisting}{listing}{listings}
\Crefname{sublstlisting}{Listing}{Listings}

\usepackage{mathrsfs}	
\usepackage{amsfonts}

\usepackage[export]{adjustbox} 

\usepackage{boldline}					

\usepackage{multirow}

\renewcommand{\paragraph}[1]{\vskip 3pt\noindent\textbf{#1 }}	 

  {\begin{list}{$\bullet$}%
     {\setlength{\parsep}{0pt}%
      \setlength{\topsep}{0pt}%
      \setlength{\itemsep}{2pt}}}%
  {\end{list}}
\newcommand\Noted[1]{} 

\definecolor{darkblue}{rgb}{0.0, 0.0, 0.55}
\definecolor{mygreen}{HTML}{ADFF2F}
\definecolor{mylightgray}{gray}{0.8}

  {\begin{itemize}
	[leftmargin=0cm,
		itemindent=.3cm,
		labelwidth=\itemindent,
		labelsep=0pt,
		parsep=0.0pt,
		topsep=0.5pt,
		itemsep=0.5pt,
		align=left]
  }%
  {\end{itemize}}    

  {\begin{enumerate}
	[leftmargin=.cm,itemindent=.5cm,labelwidth=\itemindent,
		labelsep=0pt,
		parsep=1pt,
		topsep=1pt,
		itemsep=3pt,
		align=left]
  }%
  {\end{enumerate}}    

\makeatletter
\def\@copyrightspace{\relax}
\makeatother

\begin{document}

\title{FedRDMA: Communication-Efficient Cross-Silo Federated LLM via Chunked RDMA Transmission}

\author[Zeling Zhang, Dongqi Cai, Yiran Zhang, Mengwei Xu, Shangguang Wang, Ao Zhou]{\Large{Zeling Zhang\footnotemark[1], Dongqi Cai\footnotemark[1], Yiran Zhang, Mengwei Xu, Shangguang Wang, Ao Zhou}}
\affiliation{%
  \institution{Beijing University of Posts and Telecommunications (BUPT)}
  \country{}
}


\begin{abstract}
    Communication overhead is a significant bottleneck in federated learning (FL), which has been exaggerated with the increasing size of AI models.  
    In this paper, we propose \sys, a communication-efficient cross-silo FL system that integrates RDMA into the FL communication protocol.
    To overcome the limitations of RDMA in wide-area networks (WANs), \sys divides the updated model into chunks and designs a series of optimization techniques to improve the efficiency and robustness of RDMA-based communication.
    We implement \sys atop the industrial federated learning framework and evaluate it on a real-world cross-silo FL scenario. 
    The experimental results show that \sys can achieve up to 3.8$\times$ speedup in communication efficiency compared to traditional TCP/IP-based FL systems.
\end{abstract}



\maketitle

\renewcommand{\thefootnote}{\fnsymbol{footnote}} 
\footnotetext[1]{These authors contribute equally to this work.} 

\renewcommand{\thefootnote}{\arabic{footnote}}



\section{Introduction}\label{sec:intro}

\paragraph{Cross-silo FedLLM}
Large Language Models (LLMs) have demonstrated remarkable proficiency in handling generic machine learning tasks~\cite{robinson2022leveraging, guo2023images, zhang2023sentiment, sun2023text, chae2023large}.
This advancement has led numerous companies to develop their LLMs for various applications, including computational phenotyping and medical information extraction in the medical field~\cite{zeng2018natural, goel2023llms}, as well as data mining and generation from text in the financial sector~\cite{huang2023finbert,yang2020finbert}.
Amid concerns regarding data privacy~\cite{voigt2017eu} and the challenge of isolated training data~\cite{yang2019federated, li2021survey}, federated learning (FL) has emerged as the predominant method for collaboratively training LLMs without sharing raw data, known as FedLLM~\cite{cai2023federated, cai2023efficient, xu2023federated, chen2023federated, zhao2024llm}.
This study concentrates on cross-silo FedLLM, where the clients are powerful industrial servers equipped with high bandwidth and computational resources.


\paragraph{Cross-silo FedLLM Communication Overhead}
Although acquiring more powerful GPUs can lessen computation overhead, addressing the communication overhead on WANs remains a formidable challenge, even with high bandwidth.
For instance, when federating full-tuning of the GPT-2 model (117M parameters)~\cite{radford2019language} with two NVIDIA A800 80G GPUs and 10Gbps bandwidth, it still takes 45.9s to transfer the model weights per round, accounting for more than 44.97\% of the total FL time.
This delay stems from repeated memory copies and context switchings due to the TCP protocol stack and the inherent delay of WAN itself, as will be elaborated in \S~\ref{sec:background-fedllm}.

\paragraph{Communication-Efficient RDMA}
Remote Direct Memory Access (RDMA) is a technique that has recently become widely used to reduce communication overhead in local-area network (LAN) distributed machine learning~\cite{xue2019fast, tian2021accelerating}.
It could directly transfer data between the memories of two RDMA-enabled servers, bypassing the CPU and OS kernel, which could reduce the communication overhead by up to 98.8\% as shown in \S~\ref{sec:background-rdma}.
However, RDMA requires a lossless network environment to fullfill its potential, making it nearly impractical on WANs.
This is because packet loss and delays, common on WANs, can lead to numerous retransmissions, which is unaccpectable by RDMA~\cite{guo2016rdma}, thus disminishing RDMA's performance improvemen tor even causing transmission failures.

\paragraph{Our Solution: \sys}
Inspired by the modular parameters in LLMs, we propose \sys to manage the application layer traffic flow by transferring small packets in a chunked, sequential, and smooth manner.
To further enhance performance, we design a RDMA memory pool with in-place buffering and reversed receiving strategy to effectively reassemble chunks.  
Our contributions are as follows:
\begin{itemize}
    \item We conduct preliminary experiments to demonstrate that, despite the high bandwidth and computation resources, cross-silo FedLLM still suffers from high communication overhead.
    \item We propose \sys, a communication-efficient cross-silo FedLLM system featuring chunked RDMA transmission and a series of optimizations.
    \item We implement \sys atop the industrial FedLLM framework FATE~\footnote{\url{https://github.com/FederatedAI/FATE}} and conduct extensive experiments to demonstrate it saves up to 3.8$\times$ communication time compared to TCP-based FedLLM systems.
\end{itemize}
\section{Background}\label{sec:background}
\subsection{FedLLM Communication Overhead}\label{sec:background-fedllm}
Federated learning has emerged as a prominent topic in recent research~\cite{li2020review, zhang2021survey, rieke2020future}.
It is a distributed machine learning approach that allows training locally on multiple devices without centralizing  all data on a central server.
However, federated learning's effectiveness is significantly influenced by network conditions on WAN environments~\cite{li2020federated, huang2022crosssilo}, where it faces constraints from latency and additional overhead due to network communication. 
The recent emergence of large language models~\cite{kasneci2023chatgpt, chang2023survey, wei2022emergent, chen2023federated} further exacerbates the communication burden in federated learning.
The substantial size of these models, often exceeding 100MB, poses challenges for efficient transmission on complex WAN settings.

\textbf{Communication overhead in cross-silo FedLLM.}
\begin{figure}
    \centering
        \centering
        \includegraphics[width=0.7\linewidth]{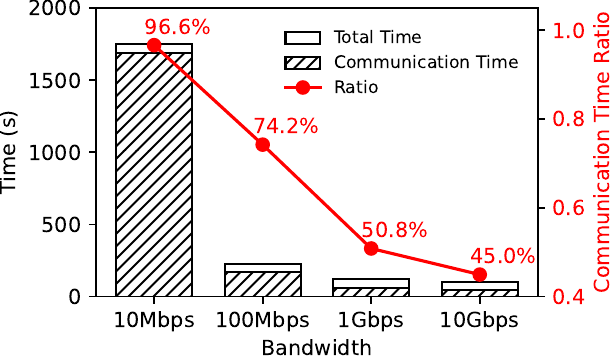}
        \label{fig:bkg-fedllm-bandwidth}
    \caption{FedLLM convergence performance under different bandwidth.}
    \label{fig:bkg-fedllm}
    \vspace{-0.2cm}
\end{figure}
In Figure \ref{fig:bkg-fedllm}, we illustrate the proportion of communication time to the overall federated learning time when full-tuning GPT2 with two NVIDIA A800 80G GPUs on a WAN environment with a 20ms round-trip time, considering different local bandwidths.
In scenario with the lowest bandwidth, the communication time significantly exceeds the computation time by 28.4$\times$.
Furthermore, it is clear that even with a high bandwidth of 10 Gbps, the communication time still constitutes a substantial 44.97\% of the total federated learning time.

Additionally, Figure \ref{fig:bkg-fedllm} reveals that in FedLLM, the model transfer performance does not scale linearly with increasing bandwidth. 
This limitation stems from several issues associated with the TCP protocol in high-bandwidth, high-performance environments~\cite{gamess2016evaluation}, as illustrated in Figure \ref{fig:bkg-rdma}. 
Challenges include the memory burden from multiple data copies, CPU overhead due to frequent context switching between user and kernel modes, and issues inherent to TCP/IP protocol design, such as slow start and complex congestion and traffic control mechanisms.

These factors introduce significant challenges for cross-silo federated training of large-scale models, which necessitates high-performance, large data transfers on complex WAN environments.

\textbf{Current solutions and their limitations.}
Some improvement methods, such as PEFT, have been proposed.
Parameter-Efficient Fine-Tuning (PEFT) is a widely recognized fine-tuning approach.     
It only fine-tunes a small fraction (typically less than 1\%) of pretrained models's parameters on new tasks data, thereby alleviating the training and communication costs of large pretrained models.
PEFT has been widely applied into FL to reduce communication overhead~\cite{cai2023efficient, cai2023federated, jiang2023low, zhao2023fedprompt, malaviya2023reducing}. 
However, PEFT typically converges slower compared to full parameter tuning, and the reduction in trainable parameters inevitably leads to a decline in model performance~\cite{qin2023federated}. 
Meanwhile, there are some other optimization methods such as backpropagation-free training paradigms~\cite{xu2023federated}, model compression/quantization~\cite{wu2018error, bernstein2018signsgd}, intelligent client selection and data sampling~\cite{li2021hermes, lipyramidfl, lai2020oort}. 
Most of them primarily target cross-device federated learning, which is not our focus.

\subsection{Communication-Efficient RDMA}\label{sec:background-rdma}
In previous studies~\cite{liu2020accelerating}, RDMA has been investigated to address the aforementioned issues.
\begin{figure}
    \centering
    \includegraphics[width=0.9\linewidth]{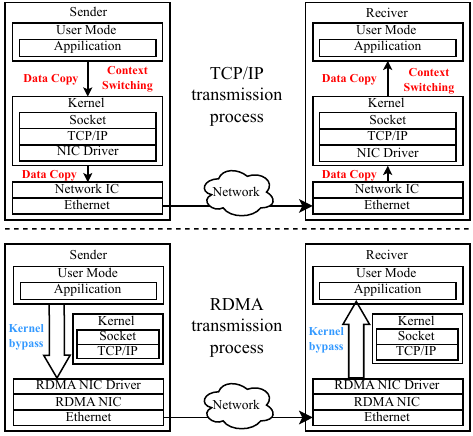}
    \caption{TCP/IP vs RDMA protocal.}
    \label{fig:bkg-rdma}
    \vspace{-0.2cm}
\end{figure}
Figure \ref{fig:bkg-rdma} highlights the differences between the RDMA and TCP protocol.
RDMA technology allows direct data transfer between the memory of two nodes, known as memory regions (MRs) in RDMA, bypassing multiple memory copies and context switchings.
This enables ultra-low latency data processing and ultra-high throughput transmission.

Despite the many advantages of RDMA, its has stringent network environment requirements. 
For instance, lossless Ethernet is usually one of the mandatory requirements~\cite{guo2016rdma} to enable RDMA, 
primarily because commercial RDMA protocols such as RoCEv2 typically support only the Go-Back-N algorithm. ~\cite{towsley1979stutter}. This results in unacceptable recovery time after packet loss, disminishing its efficiency compared to TCP.
However, constructing lossless Ethernet requires the entire network link to support key features such as Priority Flow Control (PFC)~\cite{balakrishnan1998priority} and Explicit Congestion Notification (ECN)~\cite{floyd1994tcp}, which are challenging to implement on WANs.

\textbf{Benefits and limitations of RDMA.}
\begin{figure}
    \centering
    \begin{subfigure}{0.46\linewidth}
        \centering
        \includegraphics[width=\linewidth]{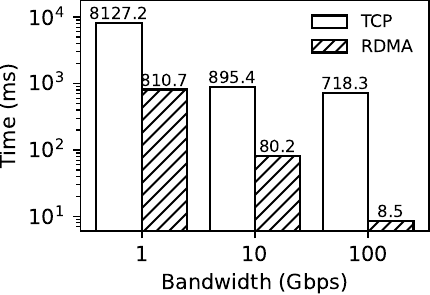}
        \caption{In-Domain (LAN)}
        \label{fig:bkg-rdma-in-domain}
    \end{subfigure}
    \begin{subfigure}{0.49\linewidth}
        \centering
        \includegraphics[width=\linewidth]{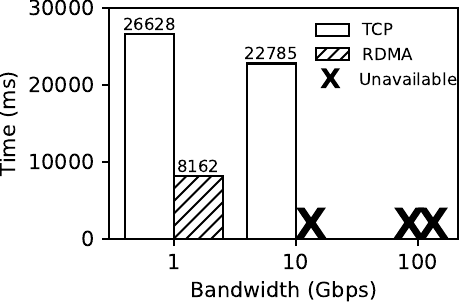}
        \caption{Cross-Domain (WAN)}
        \label{fig:bkg-rdma-cross-domain}
    \end{subfigure}
    \caption{RDMA brings significant performance improvement in-domain but fails to work cross-domain.}
    \label{fig:bkg-rdma-latency}
\end{figure}
Figure~\ref{fig:bkg-rdma-in-domain} illustrates the performance comparison between TCP and RDMA in a LAN, showing a significant performance improvement of RDMA over TCP.
However, RDMA's strict network environment requirements limit its usability on WANs. 
As Figure \ref{fig:bkg-rdma-cross-domain} shows, RDMA struggles on WANs, operating only at very low bandwidths without fully leveraging available bandwidth. 
This is because RDMA sends data too quickly, causing intermediate nodes in WAN to be overwhelmed to cache and forward a large amount of data within a short period, leading to nearly 100\% packet loss~\cite{irland1978buffer}.
Once packet is lost, RDMA steps back N packets for retransmission. 
This further increases the amount of data on WAN and triggers more packet loss, thus creating a vicious cycle that causes RDMA to fall into endless retransmissions, resulting in transmission failure.

\begin{figure*}
    \centering
    \includegraphics[width=0.8\linewidth]{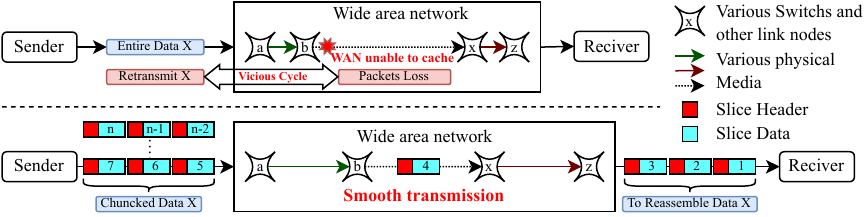}
    \caption{The process of chunking packet for smooth RDMA transfer.}
    \label{fig:design-defrdmav1}
\end{figure*}
Its rationale is that, in a single RDMA transmission scenario shown in Figure \ref{fig:design-defrdmav1}, The entire data to be transmitted is denoted as $X$. 
We can observe various switchs and link nodes in this complex WAN, each with different brands, models, and traffic-carrying capacities. 
The physical media connecting them could also vary, such as different types of fiber optics and twisted pair cables.
Therefore, when RDMA sends large data packets at once, the weak part may not be able to cache and forward, leading to a vicious cycle of packet loss and retransmission.
\section{Design} \label{sec:design}
\subsection{\sys: Chunked RDMA Transmission} \label{sec:design-truncated-rdma}
We propose \sys to address the aforementioned problem.
The main idea behind \sys is to split large data that appears in FL transmission into smaller chunks and send them sequentially to reduce the volume of data present on the link at any given time, thus controlling traffic flows in the application layer.
This helps smooth out bursty traffic and maintain stability on WAN environment. 

To do that, we use $s$ as the base chunk size to divide $X$ into $n$ smaller chunks denoted as \{$X_1$,$X_2$, $...$, $X_n$\}.
Each chunk is prefixed with a header that includes its sequence number and the total count of chunks, alongside the standard header details. 
These chunks are then sent sequentially. 
We only send the next chunk after receiving the ACK of previous one. 
This ensures that at any given time, there is at most one chunk present on the link, thus only requiring the link to be capable of trasmitting one maximum chunk size $s$.
Each chunk is temporarily stored according to the information in header in the receiver. 
Once all chunks are received, they are parsed and reassembled to complete the RDMA transmission process.

\sys enables RDMA to operate stably on WANs, with efficiency far surpassing TCP protocols. 
Moreover, it seamlessly integrates with other optimization methods such as PEFT to achieve even greater improvements as will be validated in  \S\ref{sec:evaluation-peft}. 
At the same time, \sys does not alter any algorithmic foundations and guarantees the same convergence performance and robustness as the original algorithm, which is crucial for industrial cross-silo federated learning. 

\subsection{\syse: Optimized Transmission} \label{sec:design-cross-domain-rdma}
While \sys effectively addresses the issue of RDMA being unusable on WANs, it also introduces additional overhead and latency.
Whenever splitting occurs, reassembly is required. 
This necessitates constructing a chunk header for each chunk separately at sending and parsing it when receiving. 
This introduces additional CPU overhead and increases the total amount of data transmitted. 
Moreover, when the number of chunks is relatively high, the space required to store temporary chunks imposes a substantial memory burden. 

Most importantly, since RDMA does not require CPU intervention at the receiver, sending and receiving (a.k.a, writting and reading) are asynchronous. 
To prevent the issue of data overlap caused by RDMA sending too quickly, it is necessary to introduce a artifical transmission delay to slowdown RDMA. 
In actual experiments, this delay needs to be set relatively large to ensure the robustness of the system. 
The performance degradation caused by all these additional overheads is significant and cannot be ignored.

Therefore, we have optimized \sys to address the issues encountered above, simplifying its reference as \syse.
The main idea for this optimization is to utilize pools of large receive MRs and send all chunks from back to front to the corresponding addresses to avoid additional chunk header, memory burden and further reassembly.

\begin{figure}
    \centering
    \includegraphics[width=0.9\linewidth]{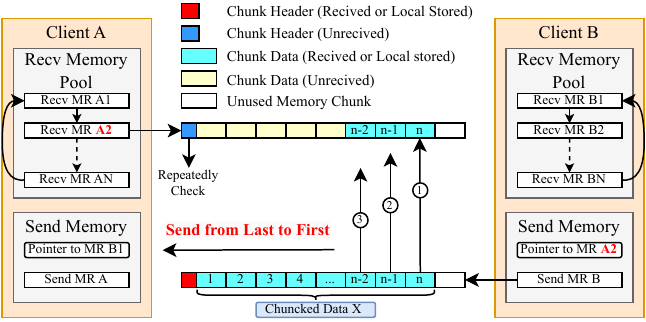}
    \caption{\syse Workflow. Client B is sending chunked data X to Client A's A2 memory region.}
    \label{fig:design-fedrdma-e}
\end{figure}
To elaborate, as shown in Figure \ref{fig:design-fedrdma-e}, in \syse, each client has one send MR and a pool of receive MRs. The capacity of these requested MRs should exceed the maximum data size of all transmissions, i.e., larger than the memory footprint of the LLM itself.
Each client maintains a pointer that points to one of the receive MR in another client.
After completing every transmission, this pointer either points to the next receive MR or loops back to the beginning.
Figure \ref{fig:design-fedrdma-e} depicts an example of \syse workflow that Client B is sending chunked data $X$ to Client A, with the pointer aimed at Receive MR A2.
At the sender Client B, the header is only attached to the first chunk.
During transmission, each chunk of data $X_n$, based on its sequential position within the entire data set $X$, is sent directly to the matching addresses in Client A's Receive MR A2, from the last chunk to the first.
Therefore, receiver can utilize the allocated receive MR directly as a temporary storage pool, and only needs to periodicly check the first few bytes of the MR.
If a valid chunk header information is found, it indicates that the sender has completed the transmission of all chunks, and the complete data is already sequentially arranged in the receive MR.

\syse reduces the header constructing and parsing operations from $n$ to 1 and brings the benefits that there is no need for extra temporary storage for further sorting and reassembly, thus significantly saving memory and CPU overhead.
Furthermore, since each chunk data $X_n$ has a unique destination address in the Receive MR, there is no conflict between sending new chunks and overwriting unread ones. 
This effectively eliminates the need for the transmission delay to slowdown RDMA, resulting in significant time savings and enhancing system performance. 

\section{Evaluation}\label{sec:evaluation}
In this section, we compare the performance of \sys with FedLLM to show its end-to-end efficiency. 
We also investigate the impact of different system hyperparameters on \sys, such as different bandwidths ranging from 1Gbps to 100Gbps and size of the best data chunk.
Afterwards, we conduct experiments to show that \sys facilitates convenient and efficient integration with PEFT methods. 
Finally,  we show \sys is more energy-efficient and environment friendly compared to traditional communication methods.

\subsection{Implementation and Setup}\label{sec:evaluation-setup}
We have developed a prototype of an domain-efficient RDMA-based federated learning communication plugin on top of the open-source industrial framework FATE. 
The implementation consists of three main components. 
Firstly, we have developed and open-sourced a user-friendly and high-performance RDMA communication library named easyPyverbs\footnote{\url{https://github.com/Marovlo/easyPyverbs}} to support further development. 
Secondly, we intercept communication traffic of the FATE framework and establish RDMA-based connection for forwarding. 
Lastly, we modify the RDMA transmission process to emulate and integrate with the gRPC streaming protocol used in the FATE framework.

\sys is derived from the aforementioned domain-efficient RDMA federated learning communication plugin design, with the additional incorporation of the \S\ref{sec:design}.

As for system cost, we deconstructed the communication process from the federated learning workflow.
Employing the same set of hyperparameters used in the aforementioned end-to-end experiment as a baseline, we conducted ten measurements and averaged the results.

\paragraph{Dataset and model}
We used the AdvertiseGen~\footnote{\url{https://www.luge.ai/\#/luge/dataDetail?id=9}} dataset to perform full-tuning on the GPT-2 model (117M parameters). 
AdvertiseGen is constructed based on the correspondence between tags on product web pages and information in the copy, representing a typical open-ended generation task. 
GPT-2 is a transformer-based LLM, trained on a dataset comprising 8 million web pages. 

\begin{table}[]
    \resizebox{\linewidth}{!}
    {
        \begin{tabular}{l|c|l|l}
            \hline
            \multicolumn{1}{c|}{\textbf{Device}} &
              \textbf{Nums} &
              \multicolumn{1}{c|}{\textbf{Device Model}} &
              \multicolumn{1}{c}{\textbf{Main Configuration}} \\ \hline 
            TOR Switch   & 2   & HUAWEI CouldEngine 6881-48s6cq & 10Gbps Ports*48, 100Gbps Ports*6  \\ \hline
            P4 Switch    & 2   & Wedge100BF-32X                 & 100Gbps Ports*32,  \\ \hline
            CORE Switch  & 2   & Inspur S6820-48XQ-AC           & 10Gbps Ports*48, 100Gbps Ports*6  \\ \hline
            RDMA NIC     & 2   & NVIDIA ConnectX-6 Dx  & 100Gbps Ports*2  \\ \hline
            Standard NIC & 2 & Intel X710 for 10 GbE SFP+     & 10Gbps Ports*2   \\ \hline
            FATE Server &
              2 &
              HREMUS 8226 &
              \begin{tabular}[c]{@{}l@{}}{NVIDIA A800 80GB},\\ Intel Xeon Gold 6226R*2,\\ 252GB DDR4 Memory\end{tabular} \\ \hline
            MININET &
              1 &
              H3C UIS 3000G5 &
              \begin{tabular}[c]{@{}l@{}}Intel Xeon Gold 5318Y*2,\\ 378GB DDR4 Memory,\\ BCM57810 10 Gigabit Ethernet*2\end{tabular} \\ \hline
            \end{tabular}
    }
    \vspace{0.2cm}
    \caption{Hardwares configurations.}
    \label{table:eval-hardwares}
    \vspace{-0.1cm}
    \end{table}
\begin{figure}
    \centering
    \includegraphics[width=0.9\linewidth]{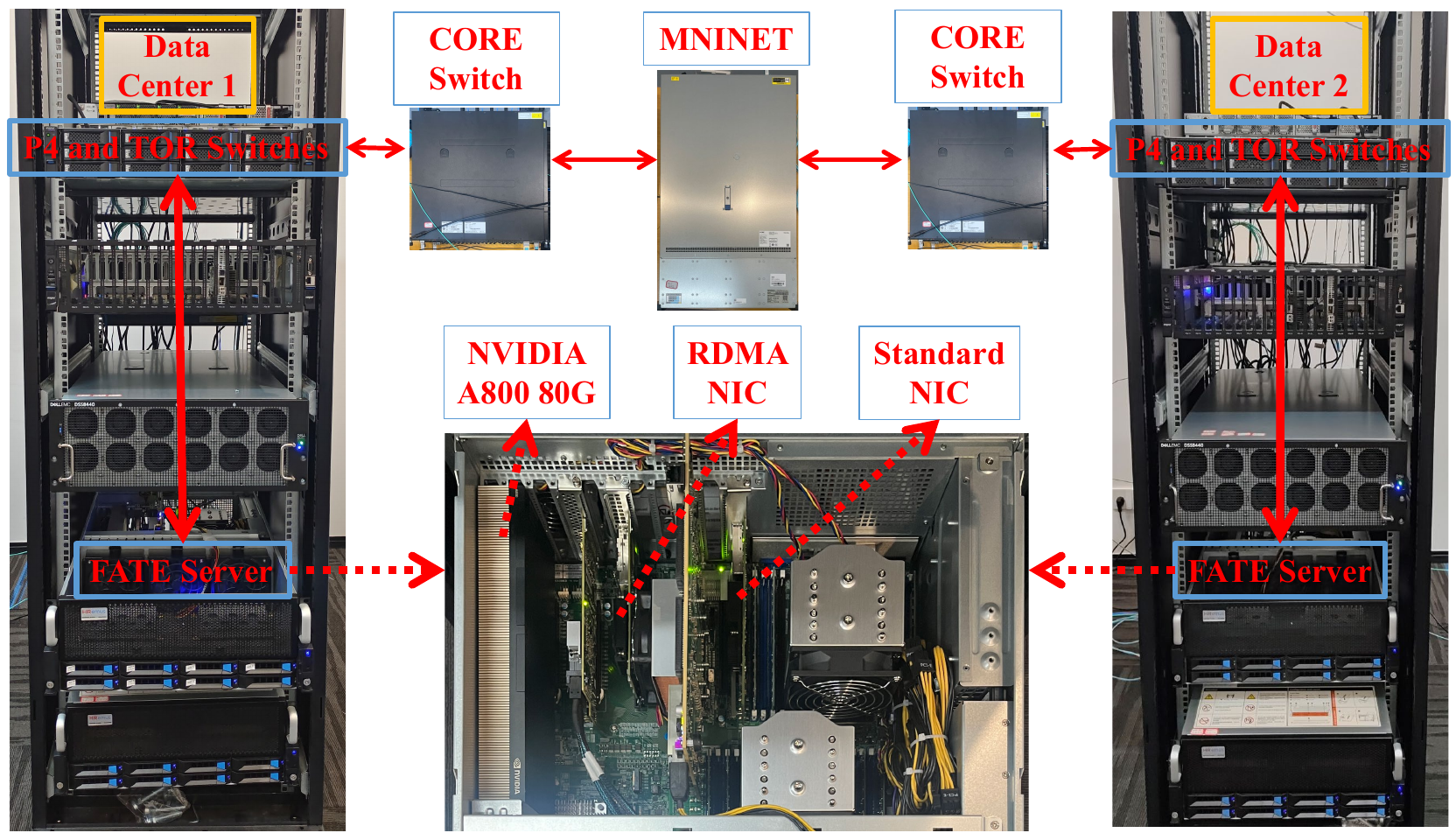}
    \caption{Physical layout and network topology.}
    \label{fig:eval-hardware-real}
    \vspace{0cm}
\end{figure}
\paragraph{Hardware and software}
We established two data centers, each hosts a FATE server with a RDMA NIC and a standard NIC.
Both data center respectively connect to two TOR switchs, which then connect to two P4 switchs, followed by two CORE switchs.
Finally, the two CORE switches connect to a MININET server that constructs two open vSwitch instances. 
The main configurations, physical layout and network topology are shown in Table \ref{table:eval-hardwares} and Fugure \ref{fig:eval-hardware-real}.
On the software side, we utilize FATE 1.11.3, FATE-LLM 1.3.0 and RDMA-CORE-37.4. 

\paragraph{Hyper-parameters}
We chose 4MB as the maximum size of a chunk in \sys.
We limit the speed of RDMA NICs to 10Gbps to align with standard NICs and introduce a 20ms RTT to all network traffic using MININET.
To simulate WAN environment, PFC and ECN on all these switches are either disabled or not supported. 

In federated learning, we employed the FedAvg~\cite{mcmahan2017communication} algorithm to conduct training for five epochs to reduce variance.
We configured the batch\_size as 32, set the learning\_rate to 5e-6, and disable secure aggregation. 

\begin{figure}
    \centering
    \begin{subfigure}{0.45\linewidth}
        \centering
        \includegraphics[width=\linewidth]{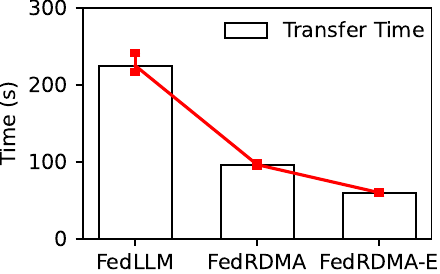}
        \caption{Communication time}
        \label{fig:eval_transfer_latency}
    \end{subfigure}
    \begin{subfigure}{0.45\linewidth}
        \centering
        \includegraphics[width=\linewidth]{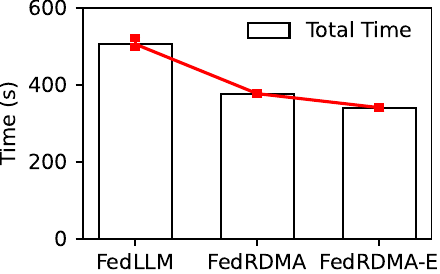}
        \caption{End to end time}
        \label{fig:eval_ete_latency}
    \end{subfigure}
    \caption{Latency comparation between FedRDMA and TCP/IP.}
    \label{fig:eval_latency}
    \vspace{-0.2cm}
\end{figure}
\subsection{End-to-end Performance}\label{sec:evaluation-end2end}
\sys outperforms TCP/IP significantly in cross-silo FedLLM. 
As depicted in Figure \ref{fig:eval_latency}, we compared \sys with FedLLM implemented by the FATE framework. 
FedLLM runs under default configurations and communicates using gRPC~\cite{wang1993grpc} based on HTTP/2, eventually derived from TCP protocol. 
During the training process, approximately 9.3GB of model weights was transmitted over WAN, constituting more than 99\% of the total communication traffic in the end-to-end training workflow.
\syse was able to reduce end-to-end communication time by 73.9\% thanks to the optimizations we made for RDMA on WANs, combined with RDMA's inherent efficiency and low overhead, as well as the drawbacks of TCP itself.
\syse ultimately result in a 33.3\% reduction in overall end-to-end federated learning time.
Moreover, even with the simpler split-merge approach used in \sys, communication efficiency was improved by 2.913$\times$, leading to an overall enhancement in federated learning efficiency by 2.381$\times$.

\begin{table}[]
  \resizebox{\linewidth}{!}
  {
      \begin{tabular}{l|l|l|l|l|l|l|l}
        \hline 
        Bandwidth (Gbps)     & 1  & 2 & 3  & 4-5      & 6-9      & 10 & 100 \\ \hline  
        Maximum chunk        & 1GB    & 1GB    & 1GB    & 12MB         & 4MB          & 4MB    & 4MB     \\ \hline 
        Best chunk           & 1GB    & 1GB    & 1GB    & 4MB          & 4MB          & 4MB    & 4MB     \\ \hline
        Link-Enable       & NO     & NO     & NO     & YES          & YES          & YES    & YES     \\ \hline 
        Latency (s)     & 8.16  & 4.10  & 2.77  & $\sim$6.57  & $\sim$6.11  & 6.00  & 5.98    \\  \hline 
        \end{tabular}
  }
  \vspace{0.2cm}
  \caption{Optimal hyperparameters under different bandwidths.
  }
  \vspace{-0.1cm}
  \label{table:eval-trans}
  \end{table}
\subsection{Impact of different hyperparameters}\label{sec:evaluation-hyperparameters}
As shown in Table \ref{table:eval-trans}, we apply various rate-limiting operations on RDMA NIC. 
Alone wtih Figure~\ref{fig:bkg-fedllm},  we can indicate that \sys continuously outperforms TCP a lot from 1Gbps to 10Gbps, and remains feasible and efficient at 100Gbps.
During the experimental process, when applying RDMA on WAN, we have some interesting observations.

We observed that at higher RDMA bandwidths, notably 4Gbps or above, transmitting large data chunks (e.g., 2MB and above) directly over the link significantly increases the risk of transmission failures.
To address this issue, it is necessary to send a smaller data chunk first, such as near the MTU size of the link.
For example, if $X_n$, the last chunk of $X$, is relatively small, it can be sent as the above mentioned small chunk, thereby optimizing one RTT time.
After this initial step, all subsequent large chunks can be smoothly transmitted. We refer to this process as Link-Enable.

We also note that when the RDMA bandwidth is small, there is no need to split the data or Link-Enable in our experimental environment.
We speculate that the amount of data simultaneously sent and present on the WAN is limited.
This does not exceed the carrying capacity of the WAN, resulting in fewer packet loss, and allowing the RDMA transmission to proceed smoothly.
This explanation also accounts for another two observations in Table \ref{table:eval-trans}: (1) \sys exhibits almost no improvement at 100Gbps compared to 10Gbps. (2) The maximum chunk size at 4-5Gbps can reach 12MB, but the best remains at 4MB.

Moreover, to validate our hypothesis, we use the Linux ``tc" command to limit the speed of the ``\textit{10Gbps} standard NIC" to \textit{10Gbps} with a token bucket filter (TBF) that can smooth TCP traffic flows.
Surprisingly, the performance of TCP actually improved, indicating that smoothing out the traffic is beneficial for both RDMA and TCP on WANs.

\begin{table}[]
  \resizebox{\linewidth}{!}
  {
      \begin{tabular}{l|l|l|l|l|l|l|l|l|l}
          \hline
          Lora Rank               & 4    & 8    & 16  & 32  & 64  & 128 & 256 & 512 & 1024 \\ \hline
          Data size (MB)  & 1.1 & 2.3 & 4.5 & 9.0   & 18.0  & 36.0 & 72.0  & 144.0 & 288.0  \\ \hline
          Num of chunks       & 1    & 2    & 4   & 5   & 7   & 12  & 21  & 39  & 75   \\ \hline
          Link-Enable       & NO   & YES  & YES & YES & YES & YES & YES & YES & YES  \\ \hline
          \end{tabular}
  }
  \vspace{0.2cm}
  \caption{The transmission data size, the total num of chunks, and whether Link-Enable is needed under different Lora Rank.}
    \label{table:eval-lora-rank}
    \end{table}
\begin{figure}
    \centering
    \includegraphics[width=\linewidth]{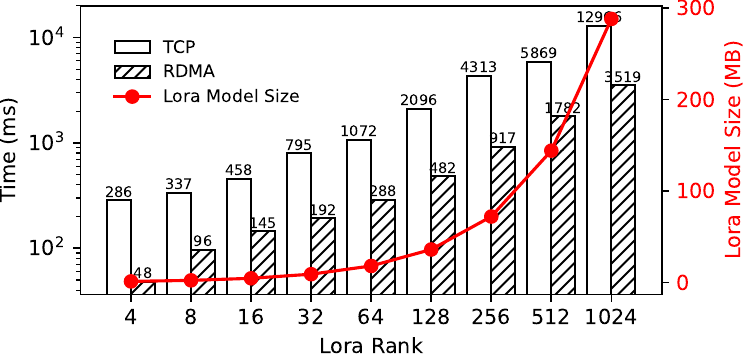}
    \caption{Communication time of FedRDMA vs TCP/IP under different PEFT settings.}
    \label{fig:eval-diff-rank-lantency}
\end{figure}

\subsection{Integration with PEFT}\label{sec:evaluation-peft}
In this section, we aim to verify the high compatibility and efficiency of \sys under different data transfer volumes introduced by varied PEFT settings.
We selected the mainstream and effective Low Rank Adaptation (LORA) method and validated a wide range of lora\_rank, the key parameter in LORA.
The lora\_rank ranges from 4 to 1024, and each case underwent five tests to reduce variance.
The sizes of transmitted data, the number of data fragments, and whether Link-Enable is required for each scenario are presented in Table \ref{table:eval-lora-rank}.

As shown in Figure \ref{fig:eval-diff-rank-lantency}, we observed that \sys reduces communication time by over 70\% in the majority of cases. 
Even in specific scenarios, such as when lora\_rank=16 a substantial improvement in communication efficiency of 2.37 $\times$ is achieved.
Therefore, \sys can complement the PEFT method well during training, and regardless of the PEFT configuration, significant improvements in communication efficiency can be achieved compared to using PEFT alone.

\begin{table}[]
  \small
  {
      \begin{tabular}{l|l|l|l|l}
        \hline
          \textbf{Method}     & \textbf{Memory}     & \textbf{Time}    & \textbf{Power} & \textbf{Energy} \\ \hline
          FedLLM     & 13.8MB        & 24.6s         & 5.1W  & 125.2J           \\ \hline
          \sys      & 60.0MB        & 9.4s          & 18.7W & 175.4J           \\ \hline
          \syse       & 0.025MB       & 6.0s          & 18.7W & 112.6J          \\ \hline
          \end{tabular}
  }
  \vspace{0.2cm}
  \caption{The system cost of transfering 1GB data.}
   \label{table:eval-system-cost}  
   \vspace{-0.2cm}
    \end{table}
\subsection{System cost}\label{sec:evaluation-system-cost}
After executing the deconstructed communication process ten times, the obtained data is presented in Table \ref{table:eval-system-cost}, where ``Memory'' refers to the remaining value after subtracting the inherent data volume of 1GB that needs to be transmitted and ``Time" refers to the total time cost transfering data.

As shown in Table \ref{table:eval-system-cost}, the refinement in memory design and optimization of the data transmission process enable \syse to achieve a 99.9\% reduction in memory overhead compared to \sys, demonstrating a significant improvement similar to that of FedLLM. 
Furthermore, despite the power of RDMA NIC being much higher than standard NIC attributed to its integration of specialized hardware for rapid data processing, the total power consumption for transmission is nevertheless reduced by more than 10\% due to the substantial decrease in transmission time.
\section{Conclusion and future work}\label{sec:conclusion}
This paper addresses a crucial and challenging issue: how to leverage RDMA technology to accelerate federated learning communication on WANs. 
We conduct detailed preliminary experiments to demonstrate that current RDMA technology fails to work effectively on WAN and the reasons behind.
To address these issues, we propose a split-reassemble-based RDMA federated learning acceleration technique, named \sys, and perform multiple optimization efforts.
We are the first to integrate RDMA into federated learning and demonstrate its effectiveness within the industrial federated learning framework.
In the future, we will extend \sys to more complex WAN environments and pave its way for large-scale cross-silo federated learning deployment.

\balance
\bibliographystyle{IEEEbib}
\bibliography{bib/ref-cdq}

\end{document}